\documentclass{article} % For LaTeX2e
\usepackage{iclr2026_conference,times}

\usepackage{amsmath,amsfonts,bm}

\def\eqref#1{equation~\ref{#1}}
\def\1{\bm{1}}

\DeclareMathAlphabet{\mathsfit}{\encodingdefault}{\sfdefault}{m}{sl}
\SetMathAlphabet{\mathsfit}{bold}{\encodingdefault}{\sfdefault}{bx}{n}

\usepackage{hyperref}
\usepackage{url}
\usepackage{graphicx} 
\usepackage{booktabs}
\usepackage{multirow}
\usepackage{caption}
\usepackage{wrapfig}

\title{Generative Frontiers: Why Evaluation Matters for Diffusion Language Models}

\author{Patrick Pynadath $^{1}$, Jiaxin Shi $^{2}$, Ruqi Zhang $^{1}$ \\
$^{1}$Department of Computer Science, Purdue University \\
West Lafayette, Indiana, 47906, USA \\
$^{2}$jiaxins.io}

\iclrfinalcopy % Uncomment for camera-ready version, but NOT for submission.
\begin{document}

\maketitle

\begin{abstract}
Diffusion language models have seen exciting recent progress, offering far more flexibility in generative trajectories than autoregressive models. 
This flexibility has motivated a growing body of research into new approaches to diffusion language modeling, which typically begins at the scale of GPT-2 small (150 million parameters). 
However, these advances introduce new issues with evaluation methodology. In this technical note, we discuss the limitations of current methodology and propose principled augmentations to ensure reliable comparisons. 
We first discuss why OpenWebText has become the standard benchmark, and why alternatives such as LM1B are inherently less meaningful. 
We then discuss the limitations of likelihood evaluations for diffusion models, and explain why relying on generative perplexity alone as a metric can lead to uninformative results. To address this, we show that generative perplexity and entropy are two components of the KL divergence to a reference distribution. 
This decomposition explains generative perplexity's sensitivity to entropy, and naturally suggests generative frontiers as a principled method for evaluating model generative quality. We conclude with empirical observations on model quality at this scale. We include a blog post with interactive content to illustrate the argument at \url{https://patrickpynadath1.github.io/blog/eval_methodology/}.
\end{abstract}

\section{Introduction}
Diffusion language models (dLLMs) offer greater flexibility in generative trajectories than autoregressive models, admitting a wide variety of training objectives, noising processes, and inference algorithms. 
This has motivated a growing body of research into new approaches to diffusion language modeling, which typically begins at the scale of GPT-2 small ($\sim$150 million parameters) — a scale that is tractable on an academic budget while still being meaningful enough to assess whether a new approach has promise.

Evaluating models at this scale, however, raises a number of methodological questions that are not always addressed explicitly in the literature. 
In this technical note, we examine these questions systematically. We discuss why OpenWebText has become the standard pretraining dataset, and why alternatives such as LM1B are less suitable for evaluating generative quality. 
We examine why likelihood-based evaluations are unreliable when comparing models with different ELBO formulations. 
We then show that generative perplexity and unigram entropy, while commonly reported together, correspond precisely to the two components of KL divergence from a reference distribution — a connection that explains why minor entropy differences can produce large perplexity shifts and cause single-point comparisons to reflect inference settings rather than model capability. 
This decomposition motivates generative frontiers as a principled evaluation framework, which we view as a principled response to methodology concerns that arise at this scale of dLLMs.

\section{Choice of Pretraining Data}
The choice of pretraining dataset has significant implications for what conclusions can be drawn from evaluation results. At the $\sim$150 million parameter scale, the dataset must be large enough to yield meaningful performance metrics, high enough quality to serve as a proxy for real language, and well-studied enough to support standard evaluation protocols. We discuss why OpenWebText has emerged as the standard choice within this regime, and why alternatives such as LM1B are less suitable.

\paragraph{Why OpenWebText is the Norm}
It is common to pretrain a $\sim$150 million parameter diffusion transformer on OpenWebText \citep{Lou_Meng_Ermon_2024, Sahoo_Arriola_Schiff_Gokaslan_Marroquin_Chiu_Rush_Kuleshov_2024, Shi_Han_Wang_Doucet_Titsias_2024, arriola2025block}. Like the original WebText, the data is obtained by scraping all outbound links on Reddit with at least 3 karma, which serves as a lightweight but effective data curation mechanism \citep{radford2021learning}. This results in a large ($\sim$9 billion tokens), diverse, yet relatively high quality corpus. For this reason, it has become the de facto pretraining set for studying language models at this scale: it sits close to the edge of what is attainable on an academic budget, while remaining similar enough to standard pretraining data to yield meaningful performance metrics. Additionally, since OpenWebText is designed as a replication of WebText — the dataset GPT-2 was trained on — it is well-studied \citep{radford2021learning, Gokaslan2019OpenWeb}, and enables the use of GPT-2 large as a reference model for generative perplexity evaluation, which we discuss further in Section 3.

\paragraph{LM1B Is Shuffled Sequence-wise}
While smaller datasets such as LM1B can be used for training language models, they differ from standard pretraining corpora in substantial ways. In particular, LM1B consists of approximately 1 billion tokens from news sources shuffled at the sentence level \citep{chelba2014billionwordbenchmarkmeasuring}, meaning there is no coherence across sentence boundaries. This was intentional: the benchmark was introduced in 2013 primarily to assess whether neural language models could outperform n-gram models at the sentence level, and was designed with likelihood-based evaluation in mind. As a result, LM1B is not well-suited as a pretraining corpus for assessing generative quality beyond the sentence level. This is consistent with how it is used in the diffusion language modeling literature, where it appears primarily in likelihood or zero-shot evaluation settings rather than as a pretraining dataset \citep{Sahoo_Arriola_Schiff_Gokaslan_Marroquin_Chiu_Rush_Kuleshov_2024, arriola2025block, sahoo2025diffusion}.

\section{Limitations of Current Methodology}
Evaluation of diffusion language models at this scale has largely followed two paradigms: likelihood-based evaluation and generative quality evaluation. While both have precedent in the autoregressive language modeling literature, both face limitations in the context of dLLMs. In this section, we discuss the limitations of each approach and motivate the need for a more principled framework.

\subsection{Different ELBOs, Different Slack to Likelihood}
It is possible to express the log likelihood as follows, for any variational distribution $q(\mathbf{z}|\mathbf{x})$:
\begin{equation}
\log p(\mathbf{x}) = \mathcal{L}(q) + D_{\mathrm{KL}}\big(q(\mathbf{z}|\mathbf{x}) \,\|\, p(\mathbf{z}|\mathbf{x})\big)
\end{equation}
where $\mathcal{L}(q) = \mathbb{E}_{q(\mathbf{z}|\mathbf{x})}\left[\log \frac{p(\mathbf{x}, \mathbf{z})}{q(\mathbf{z}|\mathbf{x})}\right]$ is the evidence lower bound (ELBO). Since $D_{\mathrm{KL}} \geq 0$, we have $\log p(\mathbf{x}) \geq \mathcal{L}(q)$, with equality if and only if $q(\mathbf{z}|\mathbf{x}) = p(\mathbf{z}|\mathbf{x})$. Crucially, the KL divergence term depends on both the true posterior --- which differs across models with different latent variable structures --- and the choice of variational family. As this gap is generally intractable to compute, direct comparison of ELBOs across models with different latent structures is unreliable for model selection \citep{bishop2006pattern}.

\paragraph{Likelihood Evaluations and Generative Quality}

Even if likelihood comparisons between different ELBO formulations were valid, they may not be relevant to the primary goal of generative quality. \citet{theis2016noteevaluationgenerativemodels} demonstrates that likelihood and generative fidelity tend to be independent of each other in high dimensions, and we refer the reader there for the full argument. It is worth noting that at larger scales, likelihood serves as a useful proxy for downstream potential after finetuning and alignment. At the GPT-2 small scale, however, there is typically no further finetuning, making it more appropriate to measure generative quality.

\subsection{Comparing Generative Quality}
\label{sec-3}
Given the difficulties with comparing ELBOs across models with different formulations, the field has moved towards generative metrics. These metrics --- exemplified by generative perplexity for coherence and unigram entropy for diversity --- are conceptually straightforward, requiring only the ability to sample from the model. However, used in isolation or at a single operating point, they can still lead to uninformative or misleading comparisons.

\begin{figure*}[t]
    \centering
    \includegraphics[width=.99\textwidth]{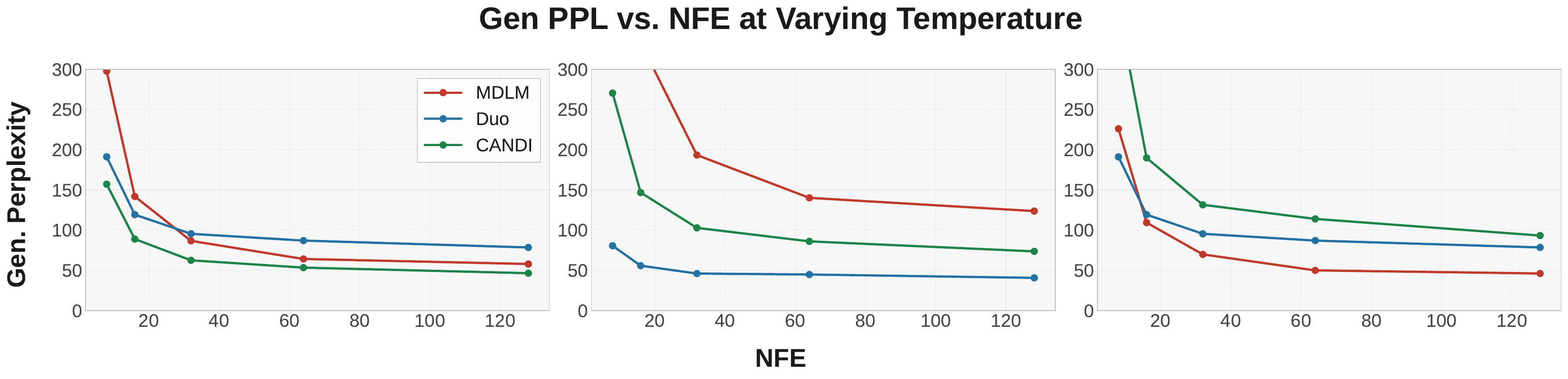}
    \caption{We show the example figure from \citet{pynadath2025candihybriddiscretecontinuousdiffusion}, where different temperature settings result in ranking reversal. This demonstrates that single point comparisons are not informative on their own, as it is always possible to achieve different rankings via minor temperature adjustments.}
    \label{fig:temp-flip}
    
    \vspace{1em}
    
    \begin{minipage}{\textwidth}
    \centering
    \captionof{table}{Sample entropy values for each method across NFE budgets under three temperature settings. As shown, all entropies for the temperature settings are reasonable -- there is no evidence of diversity hacking that leads to degenerate outputs. Yet even these minor entropy differences are sufficient to reverse rankings.}
    \label{tab:entropy_settings}
    \footnotesize
    \begin{tabular}{lccccc}
    \toprule
    \textbf{Method and Temperatures} & \textbf{NFE=8} & \textbf{NFE=16} & \textbf{NFE=32} & \textbf{NFE=64} & \textbf{NFE=128} \\
    \midrule
    \multirow{3}{*}{\shortstack[l]{MDLM=0.925 \\ Duo=1.0 \\ CANDI=0.875}}
      &  5.49   &  5.49   &  5.46   &  5.42   &  5.53 \\
      &  5.57   &  5.58   &  5.58   &  5.57   &  5.42 \\
      &  5.38   &  5.38   &  5.35   &  5.33   &  5.27 \\
    \midrule
    \multirow{3}{*}{\shortstack[l]{MDLM=1.0 \\ Duo=0.9 \\ CANDI=0.925}}
      &  5.89   &  5.79  &  5.73  &  5.69  &   5.67   \\
      &  5.07  &  5.23  &  5.25  &  5.28  &   5.28   \\
      &  5.62  &  5.59  &  5.55  &  5.58  &   5.45   \\
    \midrule
    \multirow{3}{*}{\shortstack[l]{MDLM=0.9 \\ Duo=1.0 \\ CANDI=0.95}}
      &  5.34  &  5.38  &  5.38  &  5.32  &  5.33   \\
      &  5.57  &  5.58  &  5.56  &  5.57  &  5.53   \\
      &  5.74  &  5.64  &  5.64  &  5.62  &  5.57   \\
    \bottomrule
    \end{tabular}
    \end{minipage}
\end{figure*}

\paragraph{Generative Perplexity and Excessive Repetition}
Generative perplexity is defined as follows, for a generative distribution $q_{\text{gen}}$ and a reference distribution $p_{\text{ref}}$:
\begin{align}
\text{GenPPL} &= \exp\left(\mathbb{E}_{X \sim q_{\text{gen}}}\left[\frac{1}{n}\sum_{i=1}^{n} -\log p_{\text{ref}}(x_i | x_{<i})\right]\right) \\
&= \exp\left(H(q_{\text{gen}}, p_{\text{ref}})\right).
\end{align}
This metric is popular because it is simple and relatively cheap --- all that is required is the ability to sample from $q_{\text{gen}}$ and evaluate $p_{\text{ref}}$, which can be done in a single forward pass for autoregressive models. However, generative perplexity faces a significant limitation: highly repetitive sequences can achieve competitive perplexity values while still corresponding to poor generative quality upon visual inspection \citep{zheng2024masked}. As a result, the field has moved towards also reporting unigram entropy, which captures diversity in word usage and directly addresses the failure mode of repetitive generation. Unigram entropy is computed from the empirical token frequency distribution $\hat{p}$ over a generated sequence:
\begin{equation}
\tilde{H}(q) = -\sum_{w \in \mathcal{V}} \hat{p}(w) \log \hat{p}(w).
\end{equation}
\paragraph{Generative Perplexity and Minor Entropy Changes}
While reporting unigram entropy alongside generative perplexity does meaningfully avoid extreme failure cases, \citet{pynadath2025candihybriddiscretecontinuousdiffusion} observes that even minor entropy changes are sufficient to completely reverse relative performance between methods. 
As shown in Figure~\ref{fig:temp-flip}, it is possible to change the temperature such that the relative rankings are completely different. Furthermore, there is no obvious degenerate behavior, as demonstrated by Table~\ref{tab:entropy_settings}, where we demosntrate that all methods achieve reasonable entropies within approximately 0.3 of each other. 
Yet these minor differences in unigram entropy produce entirely different rankings. 
Crucially, none of these results are misleading in isolation: each corresponds to a valid operating point obtainable by tuning the softmax temperature. 
This suggests that single-point comparisons, even when both metrics are reported, do not provide a reliable basis for model comparison.

\section{Generative Frontier Analysis}
The observations in Section~\ref{sec-3} suggest that neither likelihood-based nor single-point generative evaluations provide a reliable basis for comparing diffusion language models at this scale. 
In this section, we develop a principled framework for generative evaluation. 
We begin by showing that the metrics the community already tracks --- generative perplexity and unigram entropy --- correspond naturally to the two components of KL divergence from a reference distribution. 
This decomposition explains the sensitivity of generative perplexity to entropy, and motivates generative frontiers as a practical solution.

\subsection{KL Divergence as a Unifying Lense}
The metrics the community reports at this scale -- generative perplexity and unigram entropy -- can be naturally unified by viewing both as components of the KL divergence between the dLLM generative distribution and a reference autoregressive distribution. 
More formally:
\begin{equation}
KL(q_{\text{gen}} \| p_{\text{ref}}) = H(q_{\text{gen}}, p_{\text{ref}}) - H(q_{\text{gen}}).
\end{equation}
The first term is the cross-entropy between the generative distribution and the reference distribution, which is precisely what generative perplexity measures. 
The second term is the entropy of the generative distribution, which is approximated by unigram entropy. 
This provides a natural theoretical grounding for why both metrics are necessary, and also explains why generative perplexity is sensitive to minor entropy changes: simultaneously decreasing $H(q_{\text{gen}}, p_{\text{ref}})$ and $H(q_{\text{gen}})$ leaves the actual distance to the reference distribution unchanged, yet produces better generative perplexity. 

From the KL decomposition, we can express generative perplexity directly in terms of KL divergence and entropy:
\begin{equation}
H(q_{\text{gen}}, p_{\text{ref}}) = KL(q_{\text{gen}} \| p_{\text{ref}}) + H(q_{\text{gen}})
\end{equation}
and since generative perplexity is the exponential of the cross entropy:
\begin{equation}
\text{GenPPL}(q_{\text{gen}}) = \exp\left(KL(q_{\text{gen}} \| p_{\text{ref}}) + H(q_{\text{gen}})\right).
\end{equation}
Fixing the KL divergence and decreasing entropy by $\delta$ therefore decreases generative perplexity by a multiplicative factor of $e^\delta$. This means that a model can improve its generative perplexity simply by reducing the entropy of its generative distribution, without any improvement in the actual distance to the target distribution. Consequently, even minor entropy shifts that keep unigram entropy well within a reasonable range can produce large changes in generative perplexity, explaining the empirical observations in Figure~\ref{fig:temp-flip}.

\subsection{Moving Beyond Single Point Comparisons}
\label{single-point}
Given the relation between entropy, generative perplexity, and KL divergence, it may seem that reporting both metrics together is sufficient for reliable model comparison. 
However, we argue that single-point comparisons are insufficient for two reasons: diffusion language models parameterize an entire family of generative distributions via inference-time settings, and single-point entropy-perplexity pairs are ambiguous in terms of divergence to the target distribution.

\paragraph{LLMs Parameterize Many Distributions}
The models we wish to compare do not operate at single points --- they parameterize an entire range of generative distributions depending on the inference-time strategies used. 
For example, adjusting the temperature of the softmax directly controls the entropy of the generative distribution. 
While reducing entropy allows for potentially better generative perplexity due to the relation $\text{GenPPL}(q_{\text{gen}}) = \exp(KL(q_{\text{gen}} \| p_{\text{ref}}) + H(q_{\text{gen}}))$, it does not determine by how much the KL divergence will change, as entropy is only one component. 

There is therefore no way of knowing how a change in temperature affects KL divergence without directly measuring both metrics at that operating point. As demonstrated in Figure~\ref{fig:temp-flip} and Table~\ref{tab:entropy_settings}, tuning the temperature of each method to achieve similar entropy values is sufficient to produce entirely different rankings --- all of which are valid reflections of the respective models at those operating points.

\paragraph{Single Point Comparisons Are Ambiguous}
Even setting aside the fact that each model can be tuned to different operating points, single-point comparisons are only unambiguous in the case where one method achieves both a strictly lower generative perplexity and a strictly higher entropy than another. 
In this case, it follows directly from the KL decomposition that the first method has a lower KL divergence at that operating point. 

\begin{figure*}[t]
    \centering
    \includegraphics[width=.95\textwidth]{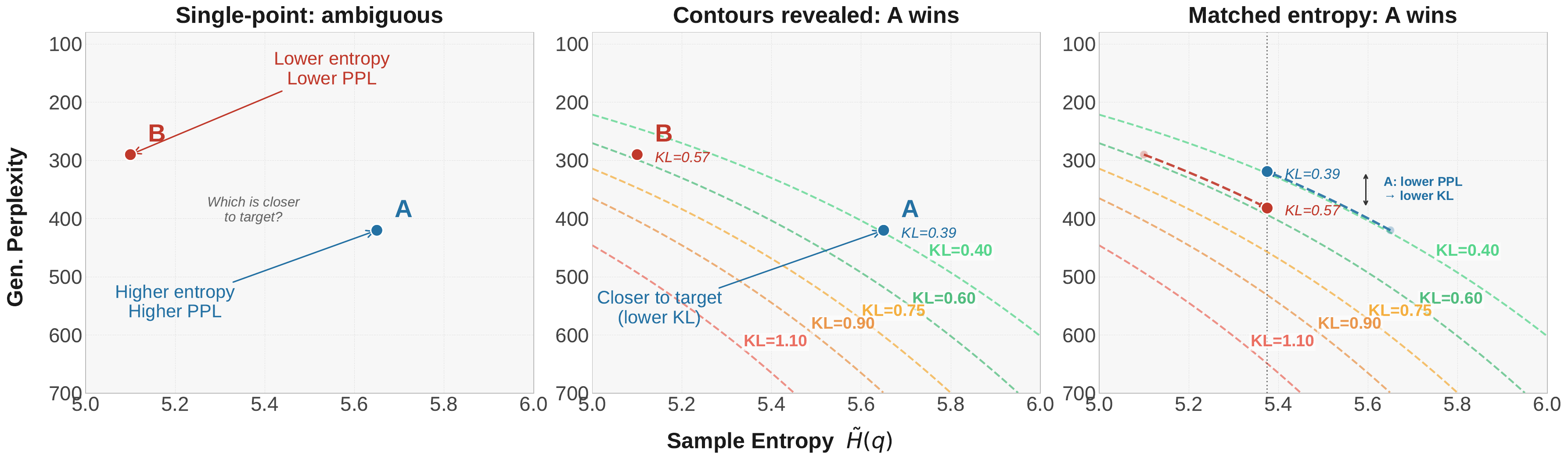}
    \caption{We demonstrate why single point evaluations are inherently ambiguous. Even though point B may have a better perplexity, plotting out the frontiers demonstrates that point A actually is closer to the target distribution. Only when points are matched on entropy or perplexity can meaningful conclusions be drawn about distance to the reference distribution.}
    \label{fig:contours}
\end{figure*}

However, strict dominance of this kind is unusual in practice --- it is more often the case that one method achieves a superior perplexity at the cost of lower entropy, or vice versa. In these cases, it is not possible to determine which method is closer to the target distribution without making assumptions about how differences in the entropy-perplexity space correspond to differences in KL divergence. As illustrated in Figure~\ref{fig:contours}, even if one point has a better perplexity, it is possible for the other to be closer in KL divergence to the target distribution, depending on the precise shape of those contours.

\subsection{Frontier Analysis as a Principled Evaluation Framework}

\begin{figure*}[t]
    \centering
    \includegraphics[width=.99\textwidth]{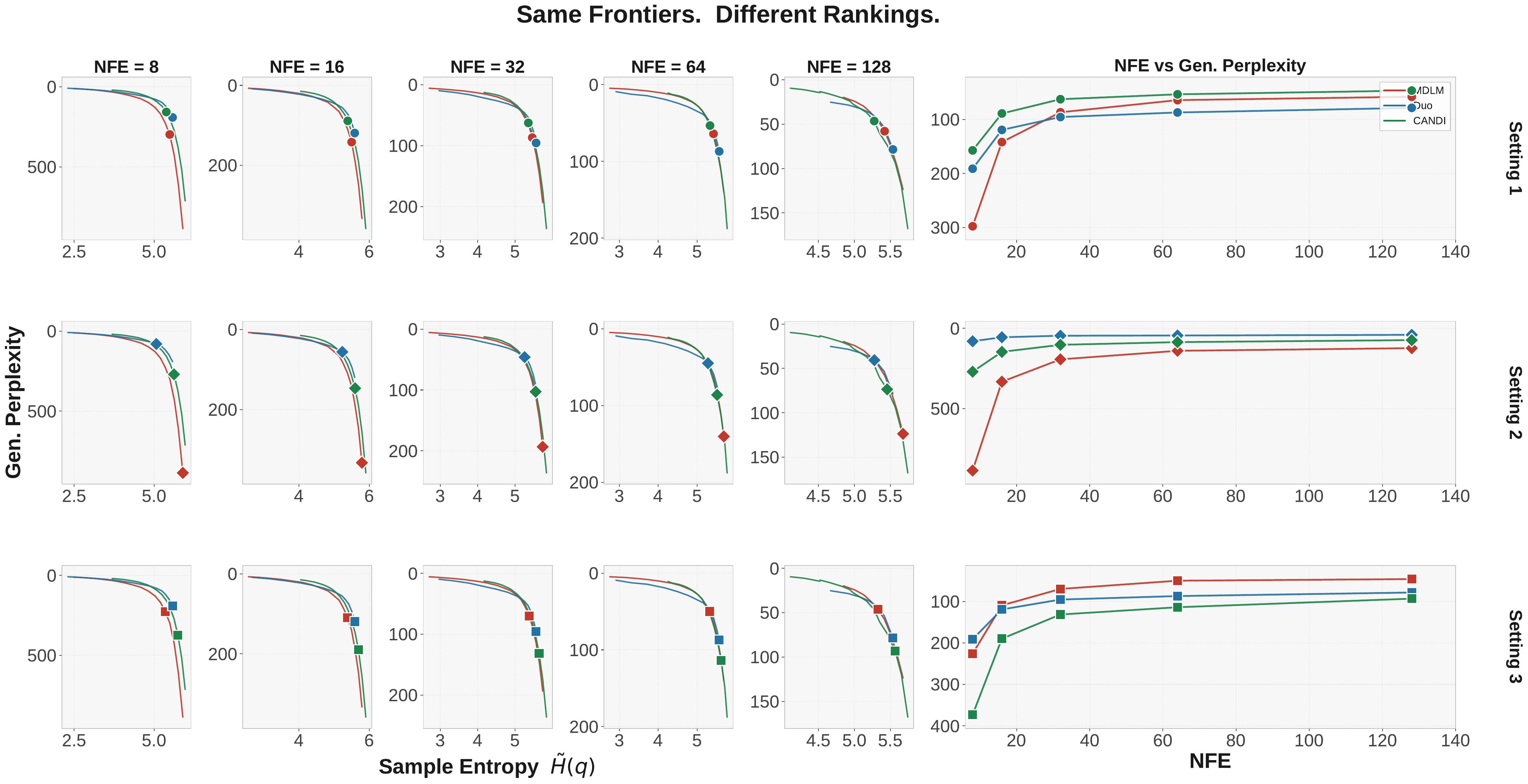}
    \caption{We visualize how single point metrics can be viewed as individual slices of the same underlying frontiers, just evaluated at different points along each curve. Thus generative frontiers provide a unifying framework for interpreting potentially disparate rankings. \textbf{While single point metrics measure inference settings, frontiers compare generative model capability.}}
    \label{fig:frontier-slices}
\end{figure*}

Generative frontiers address both limitations identified in Section~\ref{single-point} simultaneously. 
The core idea is straightforward: rather than evaluating each method at a single operating point, we sweep across a range of softmax temperatures and plot the resulting entropy-perplexity pairs as a curve for each method. 
This produces a complete picture of the generative behavior each model is capable of, and allows comparisons to be made at matched entropy or matched perplexity values rather than at arbitrary operating points. We illustrate in Figure~\ref{fig:frontier-slices} how the seemingly contradictory rankings in Figure~\ref{fig:temp-flip} are from the same generated frontier, just evaluated at different points along the curve for each method. 
We show that frontier dominance is a sufficient condition for lower KL divergence, discuss what constitutes a reasonable entropy range for comparison, and describe the assumptions under which frontier analysis is valid.

\paragraph{Dominance on Frontiers is Sufficient}
To compare methods at a fixed entropy, we fix the x-value and measure where each frontier intersects the corresponding vertical line, and likewise for generative perplexity. Under the assumption that unigram entropy comparisons accurately reflect relative rankings in joint entropy, we can make principled claims about KL divergence at each operating point. Formally, if the generative perplexities for $q_1$ and $q_2$ are equal, then by definition:
\begin{equation}
\text{GenPPL}(q_1) = \text{GenPPL}(q_2) \implies H(q_1, p_{\text{ref}}) = H(q_2, p_{\text{ref}}).
\end{equation}
If additionally $H(q_1) > H(q_2)$, then:
\begin{align}
KL(q_{1} \| p_{\text{ref}}) &= H(q_1, p_{\text{ref}}) - H(q_1) \\
&= H(q_2, p_{\text{ref}}) - H(q_1) \\
&< H(q_2, p_{\text{ref}}) - H(q_2) \\
&= KL(q_2 \| p_{\text{ref}}).
\end{align}
Thus matching generative perplexity and comparing entropy, or equivalently matching entropy and comparing generative perplexity, provides a sufficient condition for ranking methods by KL divergence. Furthermore, if one model has a strictly superior frontier than another --- lower generative perplexity at every entropy value --- it follows that it is closer to the target distribution at every operating point of interest. This is a much stronger claim than outperformance at a single operating point. Single-point comparisons measure inference settings; generative frontiers measure model quality.

\paragraph{Defining Reasonable Entropy and Perplexity Ranges} 
\begin{wrapfigure}[15]{l}{0.45\textwidth}
\vspace{-2em}
    \centering
    \includegraphics[width=0.45\textwidth]{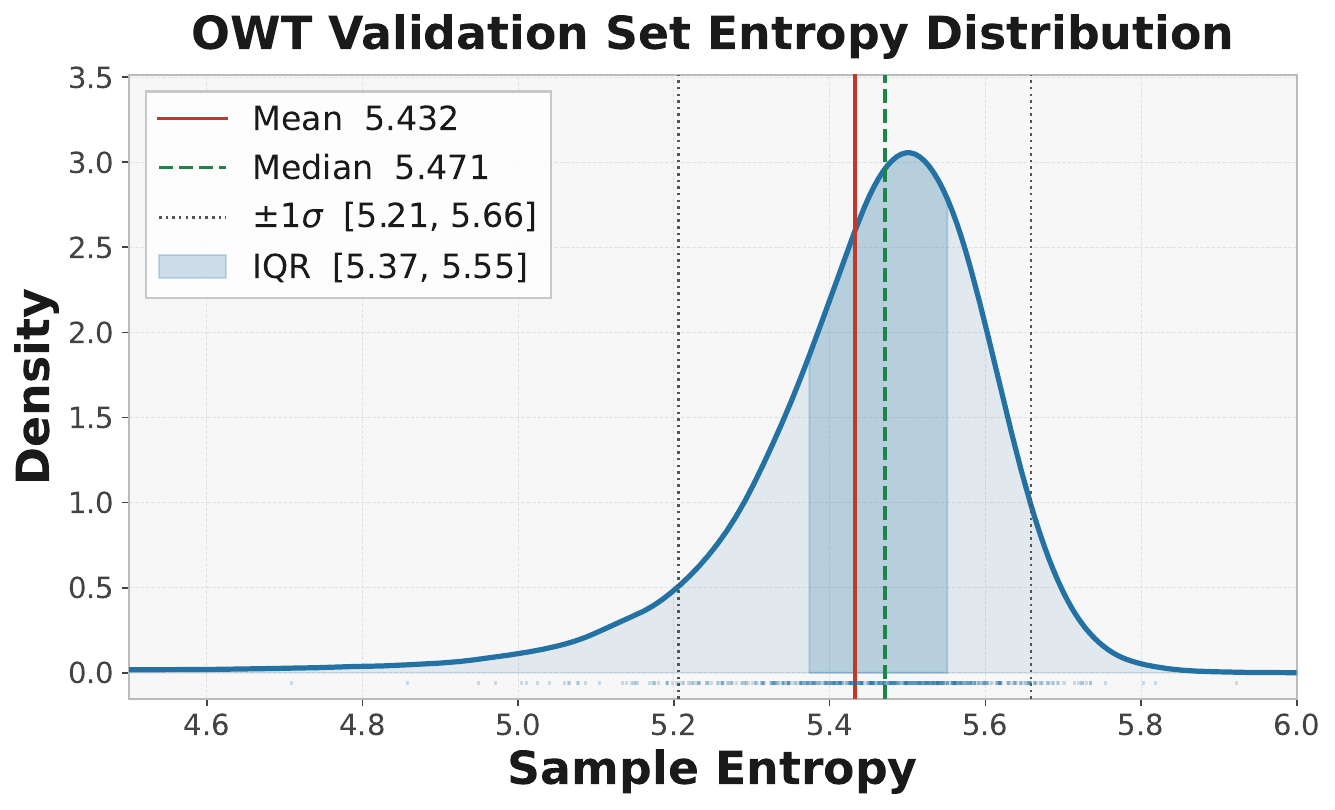}
    \caption{We show the empirical distribution of entropy values across the OpenWebText validation commonly used \citep{pynadath2025candihybriddiscretecontinuousdiffusion, sahoo2025diffusion, Sahoo_Arriola_Schiff_Gokaslan_Marroquin_Chiu_Rush_Kuleshov_2024}.}
    \label{fig:emp-entropy-dist}
\end{wrapfigure}
For generative perplexities, a natural value to look at is the evaluation perplexity of GPT-2 on the validation set. It is less clear what a reasonable entropy range is. To address this, we empirically measure the entropy distribution across the OpenWebText validation set and use it to define the range of entropies relevant for comparison. As shown in Figure~\ref{fig:emp-entropy-dist}, the mass of this distribution falls within $[5.2, 5.7]$, with a mean of $5.432$ and median of $5.471$. We use the interquartile range $[5.37, 5.55]$ as a conservative interval for identifying the most relevant operating points, and the $\pm 1\sigma$ interval $[5.21, 5.66]$ as a broader but still reasonable range. 

\begin{figure*}[t]
    \centering
    \includegraphics[width=.95\textwidth]{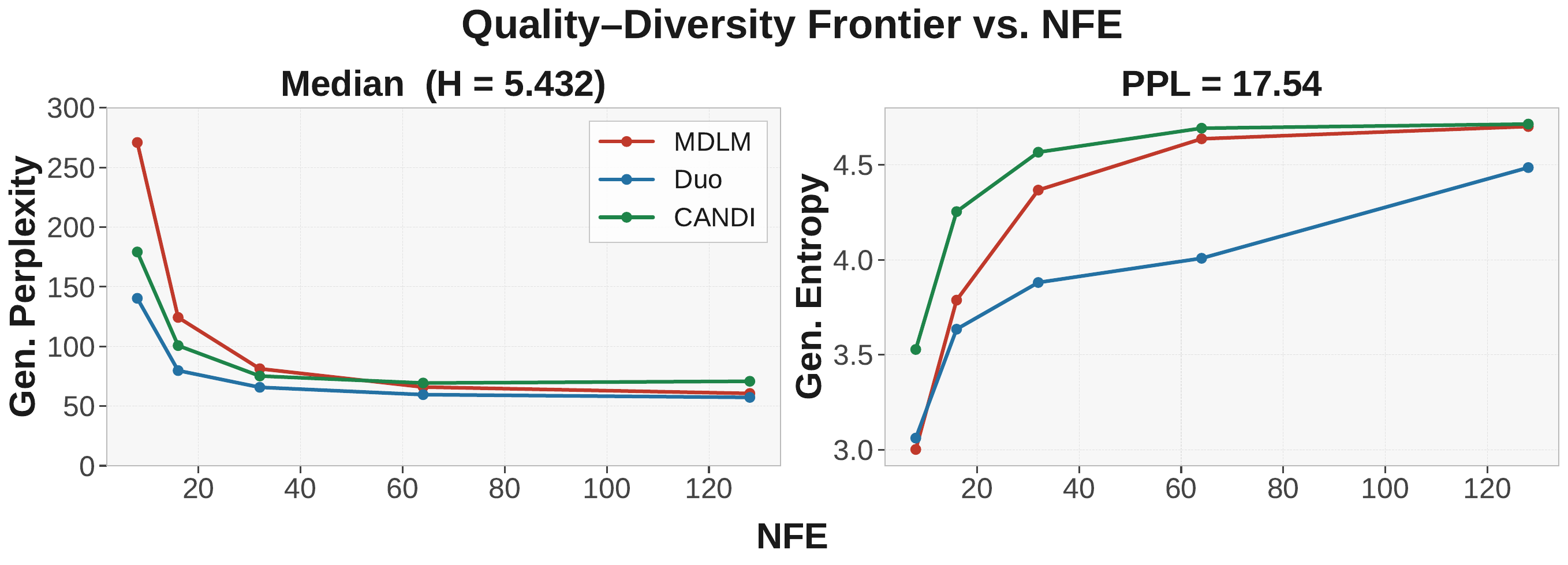}
    \caption{We illustrate how the median entropy and evaluation perplexity of the validation set of OpenWebText can be used to compare dLLMs. We use the AR Eval perplexity reported by \citet{Sahoo_Arriola_Schiff_Gokaslan_Marroquin_Chiu_Rush_Kuleshov_2024}. We observe that different models excel at different regions -- when matching the entropy of natural language, Duo excels. When matching the likelihood of natural language, CANDI excels. This demonstrates that generative quality evaluation is multi-faceted.}
    \label{fig:iqr-entropy-match}
\end{figure*}

This gives some useful guides to comparing models: given frontiers, we can fix the entropy to the empirical median and plot the resulting perplexities against NFE. Similarly, we can fix the perplexity values to the AR perplexity on OWT ($\sim$17) and compare the resulting entropies. We show an example of this in Figure~\ref{fig:iqr-entropy-match}.

\paragraph{When Frontier Analysis is Valid}
Frontier analysis relies on three assumptions. First, the reference model $p_{\text{ref}}$ must be a good proxy for the data distribution. Second, generative perplexity must accurately capture the ranking of cross entropies: $H(q, p_{\text{ref}}) < H(q', p_{\text{ref}})$ if and only if $\text{GenPPL}(q) < \text{GenPPL}(q')$. Third, unigram entropy must accurately capture the ranking of joint entropies: $H(q) < H(q')$ if and only if $\tilde{H}(q) < \tilde{H}(q')$.

The first two assumptions are straightforward to accept. GPT-2 Large is the standard reference model for generative perplexity on OpenWebText, and since the full GPT-2 was trained on the closed-source version of this dataset, it serves as a reasonable proxy distribution. Generative perplexity is an exponential of the autoregressive NLL on generated sequences, which can be viewed as a Monte Carlo estimate of the true cross entropy $H(q_{\text{gen}}, p_{\text{ref}})$.

The third assumption warrants more discussion. Unigram entropy $\tilde{H}$ is a coarser quantity than the joint entropy $H(q)$, and it is not immediately obvious that the former faithfully tracks the ranking of the latter. We argue that it is a reasonable approximation within the context of frontier analysis for the following reasons.

First, unigram entropy can be viewed as an approximation to the average marginal entropy across positions. Using $q_1, q_2, \dots, q_n$ to denote the marginal distributions at each position:
\begin{equation}
\tilde{H}(q) \approx \frac{1}{n} \sum_{i=1}^{n} H(q_i).
\end{equation}
The two quantities coincide in expectation: averaging over positions washes out position-specific idiosyncrasies, leaving only the aggregate token probabilities, which is exactly what the empirical unigram distribution captures.

Second, the sum of marginal entropies provides an upper bound on the joint entropy via the chain rule:
\begin{equation}
H(q) = \sum_{i=1}^{n} H(q_i | q_{i-1}, \dots, q_1) \leq \sum_{i=1}^{n} H(q_i) \approx n \cdot \tilde{H}(q)
\end{equation}
where the inequality follows from the fact that conditioning can only reduce entropy. This means that a smaller unigram entropy necessitates a smaller joint entropy, and a larger unigram entropy allows for a larger joint entropy --- which is the ranking property we require.

We note that the general approach of frontier analysis is entirely compatible with better approximations of $H(q)$ should they become available. We use unigram entropy here because it is already a standard metric in the field and does not require additional computation beyond what is typically reported.

\section{Empirical Observations}
\begin{figure*}[t]
    \centering
    \includegraphics[width=.95\textwidth]{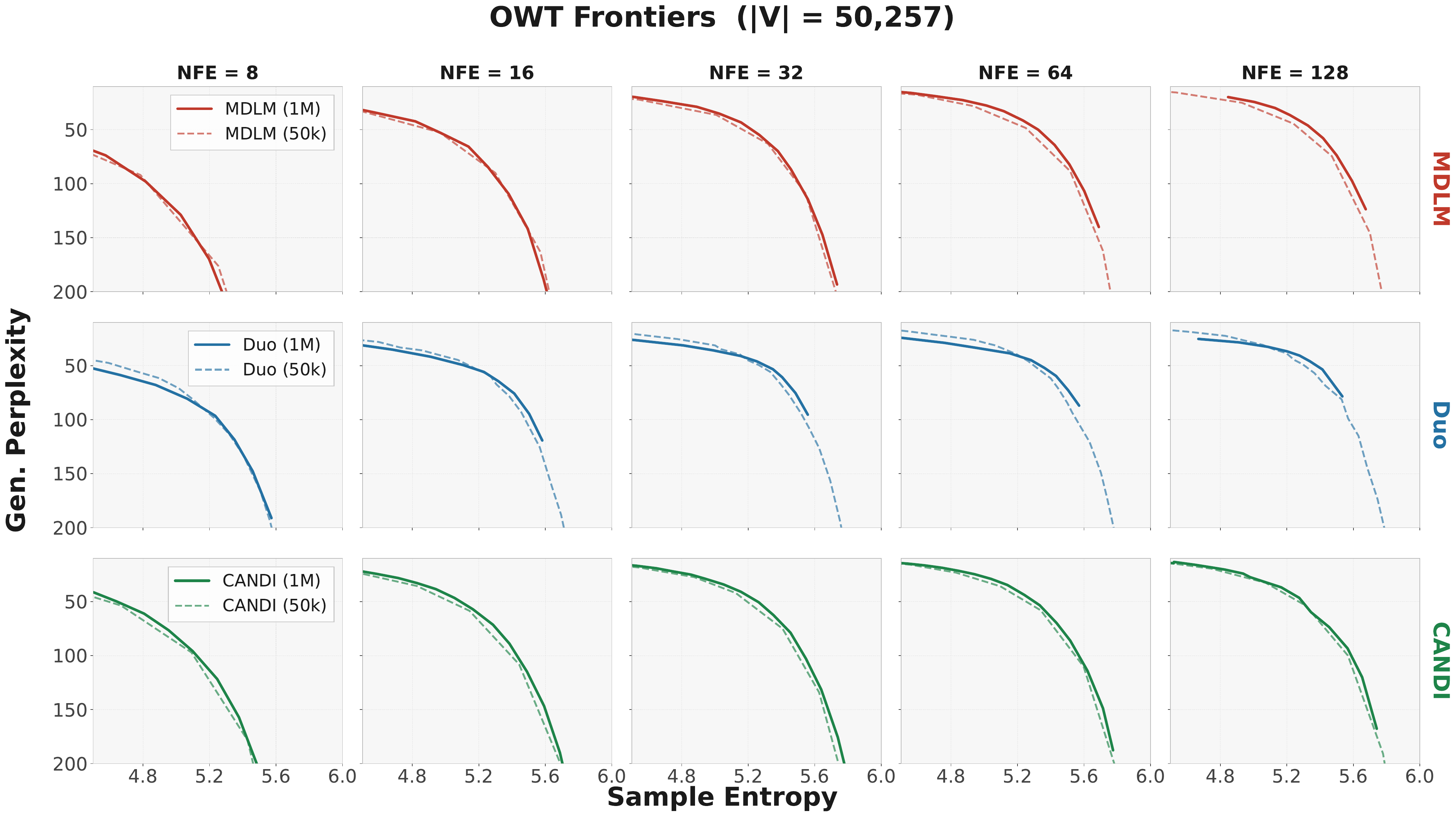}
    \caption{We compare the frontiers generated by a 50k checkpoint and a fully trained checkpoint for MDLM, Duo, and CANDI. We note that the partially trained and fully trained checkpoints align fairly closely, suggesting that 50k steps is a reasonable budget when iterating on method design.}
    \label{fig:training_behavior}
\end{figure*}

We conclude with an empirical observation that may be of practical interest to those working on pretraining diffusion language models. Using frontier analysis, we find that the generative performance of models early in training --- at 50,000 steps --- is surprisingly close to their performance at 1,000,000 steps. We visualize this in Figure~\ref{fig:training_behavior} by pretraining MDLM, DUO, and CANDI \citep{Sahoo_Arriola_Schiff_Gokaslan_Marroquin_Chiu_Rush_Kuleshov_2024, sahoo2025diffusion, pynadath2025candihybriddiscretecontinuousdiffusion} for 50,000 steps and comparing the resulting frontiers against those of fully trained checkpoints. Across all three methods, the generative frontiers at 50,000 steps closely match those at 1,000,000 steps.

This observation has a practical implication: when iterating on a new algorithm or training framework at this scale, 50,000 steps is likely sufficient to assess whether an approach has promise. If a model does not perform well by this point, it is unlikely to do so at 1,000,000 steps. While this may have already been implicitly known by practitioners who have pretrained models, the degree of similarity between the early and late frontiers is perhaps surprising.

This observation is itself a product of frontier analysis --- it would be difficult to draw this conclusion reliably from single-point comparisons alone, where differences in logit calibration across checkpoints could easily obscure the similarity. We also note a potential limitation: it is a commonly known fact that training for longer improves performance on downstream tasks, and it is not clear whether the saturation we observe here reflects a fundamental limitation of model capacity at this scale, or a limitation of generative perplexity in capturing fine-grained differences in model capability.

\section{Conclusion}
We have examined the evaluation methodology for diffusion language models at the GPT-2 small scale, identifying several pitfalls that can lead to unreliable or misleading comparisons. The key insight is that generative perplexity and unigram entropy correspond directly to the two components of KL divergence from a reference distribution, which motivates generative frontiers as a principled evaluation framework. We hope this work serves as a useful reference for researchers working on diffusion language models, and that generative frontiers become a standard part of the evaluation toolkit as the field continues to develop.

\bibliography{iclr2026_conference}
\bibliographystyle{iclr2026_conference}

\end{document}